\documentclass{article}

\PassOptionsToPackage{numbers,compress}{natbib}

\usepackage[preprint]{neurips_2026}
\workshoptitle{Agentic AI for Biological Discovery}

\usepackage[utf8]{inputenc}
\usepackage[T1]{fontenc}
\usepackage{hyperref}
\usepackage{url}
\usepackage{booktabs}
\usepackage{amsfonts}
\usepackage{amsmath}
\usepackage{amssymb}
\usepackage{nicefrac}
\usepackage{microtype}
\usepackage{xcolor}
\usepackage{graphicx}
\usepackage{subcaption}
\usepackage{tabularx}
\usepackage{tcolorbox}
\tcbuselibrary{breakable}
\usepackage{float}
\usepackage{placeins}
\usepackage[capitalize,noabbrev]{cleveref}
\usepackage{todonotes}
\usepackage{rotating}

\title{The AI Neuroscientist: An Interactive Agentic Interface for Neuroimaging Analysis}

\author{%
  Aakash Patel$^{1,2,\ast}$ \quad
  Panos Ketonis$^{1,2,\ast}$ \\
  \bfseries
  Shreya Saxena$^{2,3}$ \quad
  Smita Krishnaswamy$^{1,2,4}$ \quad
  David van Dijk$^{1,2,5}$ \\[8pt]
  \normalfont\small
  \begin{tabular}{c}
    \llap{$^{1}$}Department of Computer Science, Yale University\\
    \llap{$^{2}$}Wu Tsai Institute, Yale University\\
    \llap{$^{3}$}Department of Biomedical Engineering, Yale University\\
    \llap{$^{4}$}Department of Genetics, Yale University\\
    \llap{$^{5}$}Department of Internal Medicine, Yale University\\[4pt]
    New Haven, CT, USA\\
    \texttt{aakash.patel.ap2853@yale.edu}\\[4pt]
    \footnotesize $^{\ast}$Equal contribution.
  \end{tabular}
}

\begin{document}

\maketitle

\renewcommand{\arraystretch}{1.5}

\begin{abstract}
  Analyzing neuroimaging data requires specialized coding and statistical expertise, which limits accessibility for researchers without computational backgrounds. We present the AI Neuroscientist, a language agent for interactive data exploration. The system integrates a large language model (LLM) with a neuroimaging toolset to perform quality control, modeling, and visualization. This allows researchers to query data quality and specify analysis parameters directly in natural language, providing a transparent and interactive alternative to conventional scripted pipelines for small-scale data exploration. We demonstrate these capabilities using functional near-infrared spectroscopy (fNIRS) data, and evaluate the agent on a custom fNIRS benchmarking suite against general-purpose LLM agents with code sandboxes. Future extensions will generalize the architecture to additional modalities, including functional magnetic resonance imaging (fMRI) data, and expand the benchmarking suite to additional fNIRS tasks.
\end{abstract}

\begin{figure}[t]
  \centering
  \includegraphics[width=\linewidth]{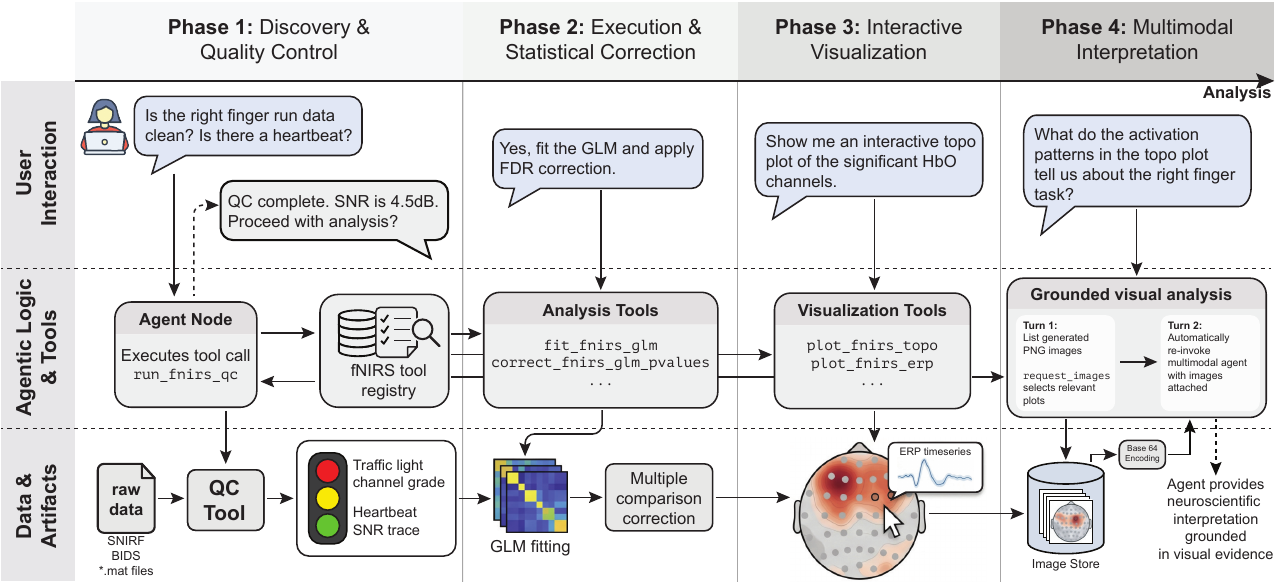}
  \caption{The Agent-Driven Neuroimaging Workflow. The workflow progresses temporally across four phases: Discovery and Quality Control, Execution and Statistical Correction, Interactive Visualization, and Interpretation. The top lane captures user queries and system feedback. The middle lane details the LangGraph-based ReAct agent logic, which routes natural language instructions to specific functions within a curated specialist tool registry. The bottom lane tracks the corresponding data transformations and generated outputs.
  }
  \label{fig:figure1_overview}
\end{figure}

\section{Introduction}

Analyzing functional neuroimaging data is a multi-step process that combines specialized libraries, file format conventions, and numerous methodological choices at each stage (e.g., selecting filter frequencies, motion artifact thresholds, and statistical models). A typical fNIRS or fMRI workflow requires navigating command-line tools, modality-specific toolboxes, and analysis scripts while judging signal quality and tuning parameters at each stage. Open standards and reproducible pipelines such as BIDS \cite{gorgolewski2016bids}, fMRIPrep \cite{esteban2019fmriprep}, and MNE-NIRS \cite{luke2021mnenirs} have made these steps more uniform, but they still assume comfort with code, documentation, and manual quality control.
Routine inspection and analysis can be inaccessible for experimentalists and clinicians without a scripting background.

Recent work shows that LLM agents can act as a layer over this kind of workflow. Agentic systems can decompose user requests, invoke external functions, and carry state across turns \cite{yao2023react, schick2023toolformer}, and similar designs have been applied to scientific domains including autonomous chemistry \cite{boiko2023autonomous, bran2024chemcrow}. A common pattern emerging in these systems is that the agent contributes most when it coordinates reliable domain tools rather than generating end-to-end answers \cite{anthropic_harness_2026, openai_harness_2026}. Neuroimaging fits this shape well: the underlying methods for preprocessing, modeling, and statistics are already well established, and what is often missing is an accessible coordination layer; one that runs them in a sensible order and surfaces the resulting evidence without requiring users to write the glue code themselves.

This paper presents the AI Neuroscientist, an agent that exposes standard neuroimaging analysis workflows through natural language, using fNIRS as an initial validation case. We use fNIRS as the first modality because the hemodynamic signal is interpretable at the channel level, files are small enough for interactive turnaround, and community guidance prescribes a clear set of checks before analysis: detection of the cardiac component in the cortical signal, identification of motion artifacts, and assessment of per-channel signal quality \cite{pinti2020present, yucel2021best}. These checks map directly onto the questions users tend to ask, such as ``is the data clean?'', ``is there a heartbeat?'', and ``which channels respond to the task?'', and they give the agent a concrete policy to follow before any inferential step.

The agent wraps a standard fNIRS analysis pipeline inside a tool registry that it calls in response to user queries. The available operations range from extracting events and segmenting the continuous data into stimulus-locked trials (epoching), to quantifying task-induced brain activation using a General Linear Model (GLM) and performing multiple-comparison correction \cite{benjamini1995controlling}. Every call writes explicit intermediate artifacts: QC dashboards, beta matrices, significance masks, and topographic plots. The user can inspect these at any point, which keeps interpretation tied to evidence instead of to a single opaque response from the model.

The intended use case is small-scale interactive exploration of one subject or a small set of runs, not unattended batch processing or clinical decision-making. Within that scope, the design illustrates a practical way to bring LLM agents into scientific workflows: constrain them to a curated set of tools, expose the assumptions baked into each step, and require human-readable artifacts at every stage so that interpretation can be checked against the data. We additionally evaluate the agent on a structured fNIRS benchmark spanning heartbeat-based quality control, motion-spike detection, and anatomical localization tasks (Section~\ref{sec:results}), and show that a specialist tool-using agent outperforms general-purpose LLM coding agents on standard fNIRS analysis tasks.

\paragraph{Contributions.}
\begin{itemize}
    \item Development of a LangGraph/LangChain AI Neuroscientist ReAct agent with a natural-language interface.
    \item Implementation of a curated neuroimaging toolset for fNIRS processing and analysis.
    \item Design of a natural-language interface that makes fNIRS analysis tools readily accessible, while preserving transparent safeguards for evidence-based interpretation.
    \item Development of an agent-agnostic fNIRS benchmarking suite and empirical comparison showing that the specialist tool-using agent outperforms general-purpose LLM coding agents on quality-control and artifact-detection tasks.
\end{itemize}

\section{Related Work}

\paragraph{Standardized neuroimaging workflows.}
BIDS and BIDS Apps provide shared data organization and containerized pipelines \cite{gorgolewski2016bids, gorgolewski2017bids}, while fMRIPrep and MNE-NIRS consolidate preprocessing into uniform routines \cite{esteban2019fmriprep,luke2021mnenirs}.
For fNIRS, SNIRF and NIRS-BIDS extend this standardization \cite{tucker2023introduction,luke2025nirs}, and consensus guidelines prescribe hemoglobin conversion, physiological checks, event-related averaging, and GLM fitting \cite{yucel2021best}.
These tools still assume scripting fluency to execute and audit each stage.

\paragraph{Modality-specific quality control.}
Automated QC differs across modalities: fMRI relies on framewise displacement and tools such as MRIQC \cite{esteban2017mriqc}, whereas fNIRS quality centers on optode--scalp coupling and intensity stability \cite{pinti2020present}.
The scalp coupling index (SCI) uses cardiac-band cross-wavelength correlation as a contact metric \cite{pollonini2014auditory}, often combined with peak spectral power (PHOEBE) and applied in sliding windows for channel pruning \cite{pollonini2016phoebe,luke2021mnenirs,yucel2021best}.
Motion correction methods such as TDDR then address spikes and baseline shifts \cite{fishburn2019temporal}.
Our agent follows this QC vocabulary---heartbeat detectability, channel grading, and user confirmation---before modeling.

\paragraph{Language agents for scientific workflows.}
Large language models can actuate scientific workflows via tool calling, as in Biomni and Claude for Life Sciences \cite{huang2025biomni,anthropic2025claude}.
Effective scientific agents further require constrained tools, state, provenance, and inspectable intermediates \cite{anthropic_harness_2026,openai_harness_2026}.
Within neuroimaging, AI work often emphasizes multimodal representation rather than transparent pipeline orchestration.
We target practical analysis access through QC-gated, artifact-first processing.

\section{Methods}

\subsection{Agent Architecture}

Figure~\ref{fig:figure1_overview} summarizes the core architecture of the AI Neuroscientist agent. The design is a staged, tool-mediated analysis loop. As shown in the middle lane of Figure~\ref{fig:figure1_overview}, the agent node routes requests into a fixed neuroscience toolset. The top lane reflects a policy-driven phase progression: discovery and quality control (QC) first, then model execution with statistical correction, then interactive visualization, and finally interpretation. Figure~\ref{fig:figure1_overview} also highlights a two-pass multimodal interpretation loop where the agent first retrieves relevant generated figures from the image store, then performs a second reasoning pass with those images attached. Finally, the bottom lane emphasizes an artifact-first state design, where each phase emits explicit outputs (QC summaries, GLM estimates, corrected statistics, and figures) and session memory is used only for short-lived orchestration state.

These architectural components were chosen to move the analysis from user intent to explicit artifacts, and then back to grounded interpretation.
The design constrains execution to concrete operations by routing requests to a fixed toolset, rather than answering from language priors alone.
The strict phase ordering serves as a scientific safeguard against common conversational-agent failure modes, such as choosing arbitrary QC thresholds or reporting significance before correction. Furthermore, the two-pass multimodal interpretation ties neuroscientific reasoning directly to visual evidence (e.g., topographies and ERP traces) rather than text-only conversation memory. This combination of policy-constrained sequencing, statistical safeguards, and visual grounding differentiates the AI Neuroscientist from a general-purpose coding agent.
A full workflow diagram detailing these processes is provided in Figure~\ref{fig:flowchart}.

\subsection{fNIRS Data Model and Assumed Preprocessing}

The agent is designed for interoperability across common fNIRS data formats, including SNIRF/BIDS intensity recordings, optical-density snapshots, and legacy concentration MAT files. Inputs are normalized into one shared recording object, so discovery, inspection, and downstream tools share a common interface regardless of provenance.

The agent uses stage-aware routing for downstream processing. Intensity inputs enter conversion and cleaning tools (optical-density transform, modified Beer--Lambert estimation \cite{cope1988system}, SCI-based channel rejection, and temporal derivative distribution repair) before modeling; optical-density inputs go directly to heartbeat and spike QC; already-converted concentration inputs skip re-conversion and proceed to second-stage analysis such as event extraction, GLM with FDR correction, and visualization. This stage-aware routing keeps hardware- and geometry-dependent choices explicit and prevents the agent from silently re-applying upstream preprocessing.

\subsection{fNIRS Toolset}

The agent utilizes a specialized suite of fNIRS tools designed to handle the core stages of the neuroimaging pipeline, with a full itemized list and implementation details provided in Appendix~\ref{appendix:toolset}.
The tools are organized into six primary functional groups:
\begin{enumerate}
    \item \textbf{Discovery and Inspection:} Locate available recordings and summarize session metadata before analysis.
    \item \textbf{Quality Control:} Assess signal quality and flag unusable channels or subjects.
    \item \textbf{Conversion and Cleaning:} Prepare and convert raw intensity into haemodynamic signals.
    \item \textbf{Modeling and Statistics:} Fit task contrasts and quantify statistical significance.
    \item \textbf{Visualization and Interpretation:} Generate and retrieve figures for multimodal result inspection.
    \item \textbf{Literature Search and Review:} Query and summarize related neuroscientific literature.
\end{enumerate}

A session-scoped memory system, the \textit{image store}, manages the interpretation workflow. As figures are generated or discovered, they can be registered and retrieved with descriptive metadata. For users seeking neuroscientific validation, the agent utilizes these tools to attach images for a second reasoning pass. This ensures that interpretation of brain activation is grounded in the underlying visual evidence.

\section{Benchmark Design}
\label{sec:benchmark_design}

We evaluate agents on a custom fNIRS benchmarking suite (\emph{fnirs\_bench}). The suite is agent-agnostic: any system that writes the declared artifacts under a natural-language prompt can be scored. Full task definitions, data sources, and qualitative error analyses are provided in Appendix~\ref{appendix:benchmark_suite}.

\subsection{Design rationale}

The benchmark targets three capabilities that recur in interactive fNIRS workflows and that are difficult for general-purpose coding agents to reproduce reliably:
\begin{enumerate}
    \item \textbf{Physiology-aware quality control.} Optode--scalp coupling is often assessed via a cardiac ($\sim$1\,Hz) spectral peak in optical density (OD). Agents must translate this criterion into a per-channel mask and a subject-level usability decision (Tasks~1a--c).
    \item \textbf{Artifact localization.} Motion and other events appear as amplitude spikes in OD time series. Agents must output a boolean \texttt{spike\_mask} aligned to channel names (Tasks~2a--b).
    \item \textbf{Anatomical plausibility of activation maps.} Interpretation requires judging whether a topographic pattern matches the expected response for a given task (Task~3).
\end{enumerate}
Tasks 1 and 2 include an \emph{easy} synthetic variant (controlled ground truth from a shared MNE motor template) and a \emph{hard} real recording drawn from OpenNeuro BIDS exports, so failures reflect both algorithmic mistakes and brittleness on messy data.

\subsection{Harness, agents, and run protocol}

The evaluation harness presents each agent with a natural-language task prompt and staged inputs, runs the agent in a fresh working environment, collects the declared output artifacts, and scores them against held-out ground truth. Agents that execute arbitrary code run inside containers that mount only the task inputs (read-only) and a writable output directory, with no access to reference implementations, ground-truth labels, or artifacts from other runs. Per-run logs record tool trajectories, text answers, token usage, and base scores for later reliability analysis.

We compare three agents over five independent runs per task (\texttt{n\_runs=5}):
\begin{itemize}
    \item \textbf{AI Neuroscientist} --- the \texttt{fnirs\_agent} specialist with curated path-based tools (heartbeat SNR, MAD spikes, SNIRF I/O, topomap plotting, etc.) and \textbf{no} sandbox code execution by default; backbone \texttt{gpt-5-nano}.
    \item \textbf{LLM (gpt-5-nano)} --- general-purpose model in an isolated Docker sandbox (\texttt{run\_python}, \texttt{bash}, file I/O only); same backbone as the specialist agent.
    \item \textbf{LLM (claude-sonnet-5)} --- same sandbox stack with a stronger model.
\end{itemize}
The specialist agent receives staged NPZ/SNIRF/PNG inputs and may call registered tools directly. Vision tasks attach PNG stimuli on the user message; numeric tasks require NPZ artifacts with a \texttt{ch\_names} array so channels align by name rather than row order.

\subsection{Scoring framework}

Each task run yields a \textbf{base score} $b \in [0,1]$ from task-specific metrics. With $N$ repeated runs, the task base score is $\bar{b} = \frac{1}{N}\sum_{k}b_k$.

\textbf{Reliability penalties} discourage unstable tool use and unstable numerical artifacts across runs:
\begin{itemize}
    \item \textbf{Tool-trace consistency} $S_{\mathrm{trace}} \in [0,1]$: mean pairwise Jaccard similarity of the sets of tools invoked across runs ($1.0$ = identical tool usage every time).
    \item \textbf{Matrix variance} $V_{\mathrm{matrix}}$: maximum element-wise variance of the task's primary numerical artifact (e.g.\ \texttt{good\_mask}, \texttt{spike\_mask}) across runs; $0$ for text-only tasks or identical outputs.
\end{itemize}
With defaults \texttt{trace\_tol=1.0}, \texttt{trace\_penalty\_weight=0.5}, \texttt{trace\_scale=1.0}, \texttt{var\_tol=$10^{-6}$}, \texttt{variance\_penalty\_weight=0.5}, \texttt{variance\_scale=0.25}:
\begin{align*}
    p_{\mathrm{trace}} &= 0.5 \cdot \min\!\left(1,\; \frac{\max(0,\; 1 - S_{\mathrm{trace}})}{1.0}\right), \\
    p_{\mathrm{var}} &= 0.5 \cdot \min\!\left(1,\; \frac{\max(0,\; V_{\mathrm{matrix}} - 10^{-6})}{0.25}\right), \\
    s_{\mathrm{final}} &= \max(0,\; \bar{b} - p_{\mathrm{trace}} - p_{\mathrm{var}}).
\end{align*}
The overall score is the unweighted mean of per-task $s_{\mathrm{final}}$ values. Table~\ref{tab:benchmark-scores} reports $s_{\mathrm{final}}$ per task with standard error computed from the per-run base scores.

\section{Benchmark Results}
\label{sec:results}

\begin{table}[t]
    \centering
    \small
    \setlength{\tabcolsep}{2.2pt}
    \begin{tabular}{lccccccc}
\toprule
Agent & T1a & T1b & T1c & T2a & T2b & T3 & Overall \\
\midrule
AI Neuroscientist & \textbf{0.90$\pm$0.00} & \textbf{0.85$\pm$0.00} & \textbf{0.90$\pm$0.00} & 0.67$\pm$0.20 & \textbf{0.90$\pm$0.00} & 0.50$\pm$0.00 & \textbf{0.79$\pm$0.05} \\
LLM (claude-sonnet-5) & 0.80$\pm$0.00 & 0.65$\pm$0.00 & 0.00$\pm$0.00 & \textbf{0.77$\pm$0.00} & 0.00$\pm$0.17 & 0.50$\pm$0.06 & 0.45$\pm$0.07 \\
LLM (gpt-5-nano) & 0.19$\pm$0.06 & 0.14$\pm$0.02 & 0.00$\pm$0.02 & 0.00$\pm$0.00 & 0.00$\pm$0.00 & 0.50$\pm$0.00 & 0.14$\pm$0.06 \\
\bottomrule
\end{tabular}

    \caption{Benchmark scores by task ($\pm$ SEM). Scores are aggregated across five runs per agent. T1a--c: heartbeat QC (synthetic; real mixed; real poor SNR). T2a--b: spike detection (synthetic; real). T3: anatomical localization. Overall is the mean across tasks.}
    \label{tab:benchmark-scores}
\end{table}

Table~\ref{tab:benchmark-scores} summarizes final scores on the six-task suite. The AI Neuroscientist achieves the highest overall score ($0.79\pm0.05$), substantially outperforming GPT ($0.14\pm0.06$) and the stronger Claude agent ($0.45\pm0.07$). Gains are largest on real and poor-SNR quality-control and spike-detection tasks (T1c, T2b), where both sandbox LLMs score near zero while the AI Neuroscientist remains near $0.9$. On anatomical localization (T3), all three agents perform similarly ($\approx$0.5), indicating that tool specialization helps most when the required computation maps onto curated signal-processing tools rather than open-ended visual reasoning. Figure~\ref{fig:deepdive-t1} illustrates failure modes on heartbeat QC; spike detection and anatomical localization are analyzed in Appendix~\ref{appendix:benchmark_suite} (Figures~\ref{fig:deepdive-t2}–\ref{fig:deepdive-t3}).

\begin{figure}[t]
    \centering
    \includegraphics[width=\linewidth]{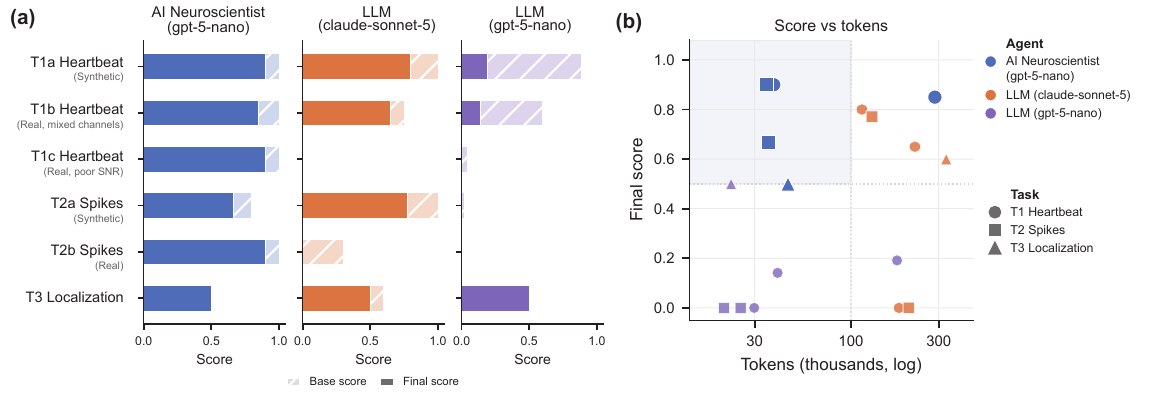}
    \caption{Benchmark diagnostics. (a)~Base artifact score versus final score after the reliability penalty. (b)~Final score versus total tokens (log scale).}
    \label{fig:benchmark-main}
\end{figure}

\begin{figure}[t]
    \centering
    \includegraphics[width=\linewidth]{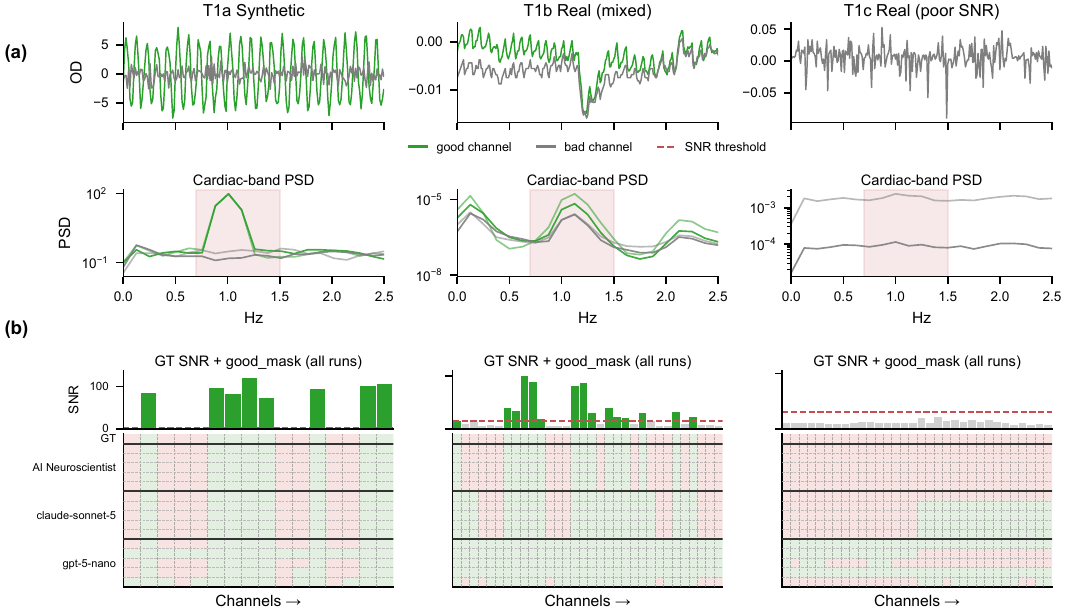}
    \caption{Task~1 deep dive: heartbeat-based QC. (a)~Example OD traces and cardiac-band PSDs for T1a (synthetic), T1b (real mixed), and T1c (real poor SNR). (b)~GT SNR bars and \texttt{good\_mask} heatmaps (GT + all agent runs). Mean $F_1$ per agent annotated; solid lines separate agents.}
    \label{fig:deepdive-t1}
\end{figure}

Figure~\ref{fig:benchmark-main} illustrates the primary advantages of the AI Neuroscientist: consistency under reliability-aware scoring and computational efficiency. Panel~(a) contrasts base artifact scores with final scores post-reliability penalty. Across T1 and T2, sandbox baseline LLMs frequently achieve moderate base scores but suffer heavy trajectory penalties from inconsistent tool execution, substantially lowering their final scores. By contrast, the AI Neuroscientist retains consistently high final scores with minimal penalties due to deterministic tool invocation paths across runs (detailed penalty compositions appear in Figure~\ref{fig:app-reliability}). Panel~(b) demonstrates token efficiency by plotting final score against total tokens on a logarithmic scale. The specialized agent achieves higher performance at significantly lower token budgets, whereas generalist sandboxes like Claude expend substantial token overhead repeatedly re-implementing analytical routines in Python. Complete runtime and token distributions per task are detailed in Figure~\ref{fig:app-cost}. These results highlight that curated tool integration enhances both task success and resource utilization relative to unconstrained code generation.

\paragraph{Heartbeat QC.}
Figure~\ref{fig:deepdive-t1} shows representative inputs, expert ground truth, and agent runs for Tasks~1a--c. Panel~(a) shows example OD traces and cardiac-band power spectral densities for one good and one bad channel in each variant (T1a synthetic, T1b real mixed, T1c real poor SNR). In T1a, the good channel exhibits a sharp $\sim$1\,Hz peak above the dashed SNR threshold; the bad channel is flat. T1b mixes both behaviors on the same montage. T1c is the stress test: even ``good-looking'' waveforms lack a cardiac peak---the correct scientific conclusion is that \emph{no} channels pass and the subject is not usable.

Panel~(b) stacks per-channel GT SNR bars (green = good) with a heatmap of \texttt{good\_mask} rows: GT followed by every run for each agent. On T1a, all agents often agree with GT. On T1b, the AI Neuroscientist matches GT across runs ($F_1 \approx 0.85$), while sandbox models omit good channels or flip borderline calls. On T1c, the dominant sandbox failure is marking channels good despite absent cardiac SNR (false positives); the specialist agent correctly outputs an all-false mask. High \texttt{good\_mask} matrix variance, stemming from variable Python scripts each run, contributes additional penalty beyond the low $F_1$.

\section{Case Study: End-to-End Analysis Demonstration}
\label{sec:case_study}

The AI Neuroscientist is designed to translate high-level user intent into a reproducible neuroscientific workflow. As a complementary qualitative demonstration, we present a case study of an end-to-end interactive analysis session on a representative fNIRS dataset (see Appendix~\ref{appendix:full_demo} for the full session transcript). The following sections highlight the agent's behavior across discovery, quality control, execution, and interpretation.

\subsection{Data Discovery and Inspection}
When provided with a raw data directory, the agent autonomously maps file identifiers to experimental conditions (e.g., \texttt{Face}, \texttt{FingerR}). When prompted (``What fNIRS runs are available in the data directory?''), the agent invokes \texttt{list\_fnirs\_runs} to index the files and returns a structured list of available runs. It then inspects the selected run (\texttt{load\_fnirs\_data}), extracting key metadata such as channel count, sampling rate, signal ranges, and event onsets, providing the user with a comprehensive overview before any processing begins.

\subsection{Interactive Quality Control and Threshold Adjustment}
A core feature of the agent is its ability to enforce data quality standards while accommodating user feedback. When instructed to evaluate data cleanliness, the agent executes quality-control tools such as heartbeat SNR and channel grading. In our demonstration, the agent initially flagged a run with a ``RED'' overall grade due to low heartbeat signal-to-noise ratio (median SNR $< 0$ dB) across most channels.

Using its multimodal visual grounding loop, the agent retrieved the generated QC topography image and provided a spatial interpretation, noting that the issue was system-wide rather than localized. The user then prompted the agent to relax the strict default thresholds (``Can you re-run QC with a heartbeat SNR threshold of -2 dB for green?''). The agent adjusted the tool parameters, re-executed the QC pipeline, and provided an updated interpretation, confirming that the data was usable for exploratory analysis while advising the exclusion of four specific channels.

\subsection{End-to-End Execution and Statistical Safeguards}
Once the data is cleared for analysis, the agent executes the modeling pipeline. Instructed to ``run the analysis pipeline,'' the agent chains event extraction, epoching, and GLM fitting. The agent automatically chooses to apply False Discovery Rate (FDR) correction to the GLM beta estimates before generating final topographies or making any claims about significant activation. This acts as a built-in scientific safeguard against reporting uncorrected p-values.

\subsection{Multimodal Interpretation of Results}
The final phase of the workflow demonstrates the agent's capacity for visually-grounded reasoning. Following the pipeline execution, the agent generates interactive HTML plots and static PNG topographies. When asked to interpret the results, the agent utilizes the \texttt{list\_image\_store} and \texttt{request\_images} tools to retrieve the topographic map of significant HbO activation. It then performs a second reasoning pass on the visual evidence, describing the spatial pattern of the activation (e.g., localized motor cortex response for a finger-tapping task) and verifying the physiological plausibility by checking the concordance of HbO increases with HbR decreases in the generated ERP traces (Figure~\ref{fig:demo-erp}).

\begin{figure}[t]
    \centering
    \begin{subfigure}{0.48\textwidth}
        \centering
        \includegraphics[width=\linewidth]{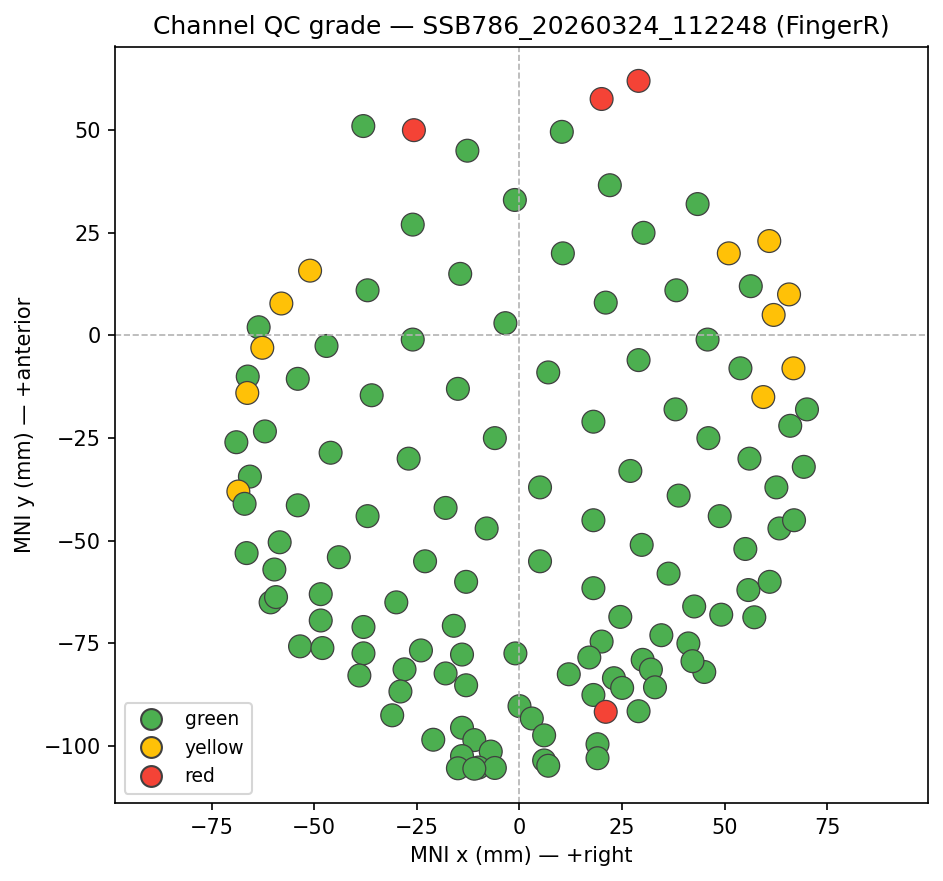}
        \caption{QC Topography}
    \end{subfigure}\hfill
    \begin{subfigure}{0.48\textwidth}
        \centering
        \includegraphics[width=\linewidth]{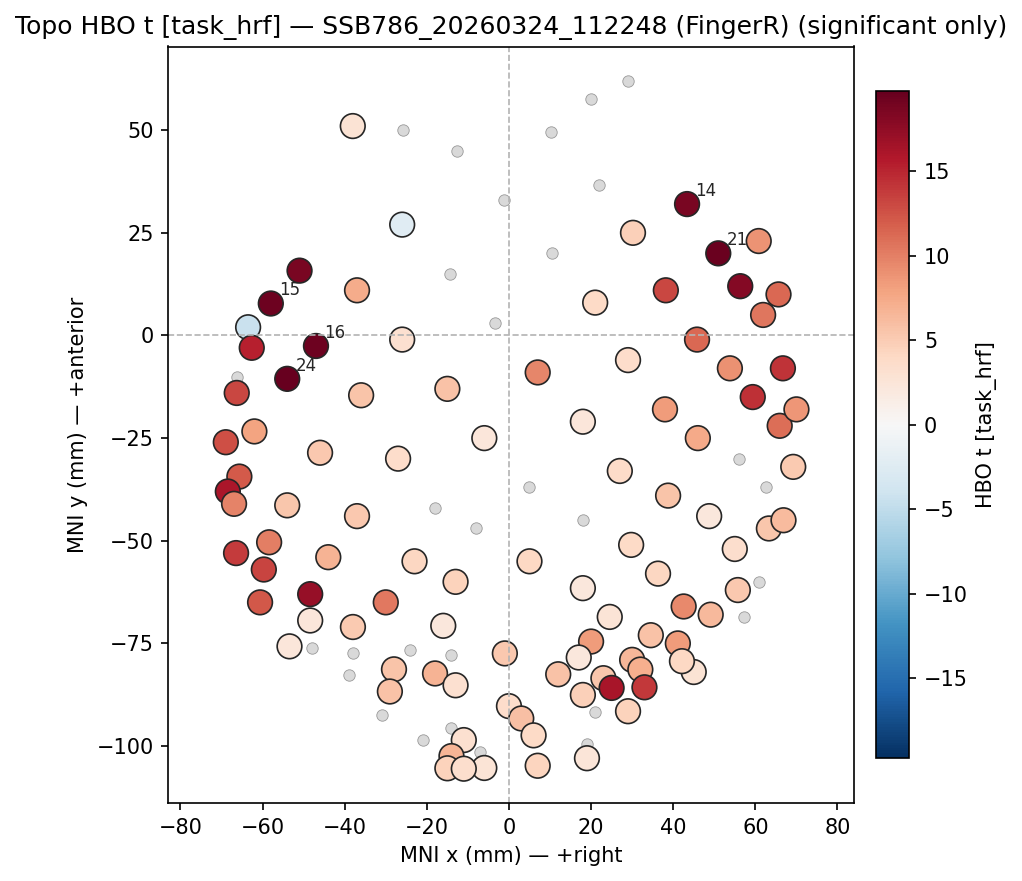}
        \caption{Significant Activation}
    \end{subfigure}
    \caption{Representative artifacts generated by the AI Neuroscientist during the interactive session. (a) The quality control topography showing channel grades based on heartbeat SNR and motion metrics. Red channels are recommended for exclusion. (b) The final topography of significant oxygenated hemoglobin (HbO) activation after FDR correction.}
    \label{fig:demo-output}
\end{figure}

\begin{figure}[t]
    \centering
    \includegraphics[width=\linewidth]{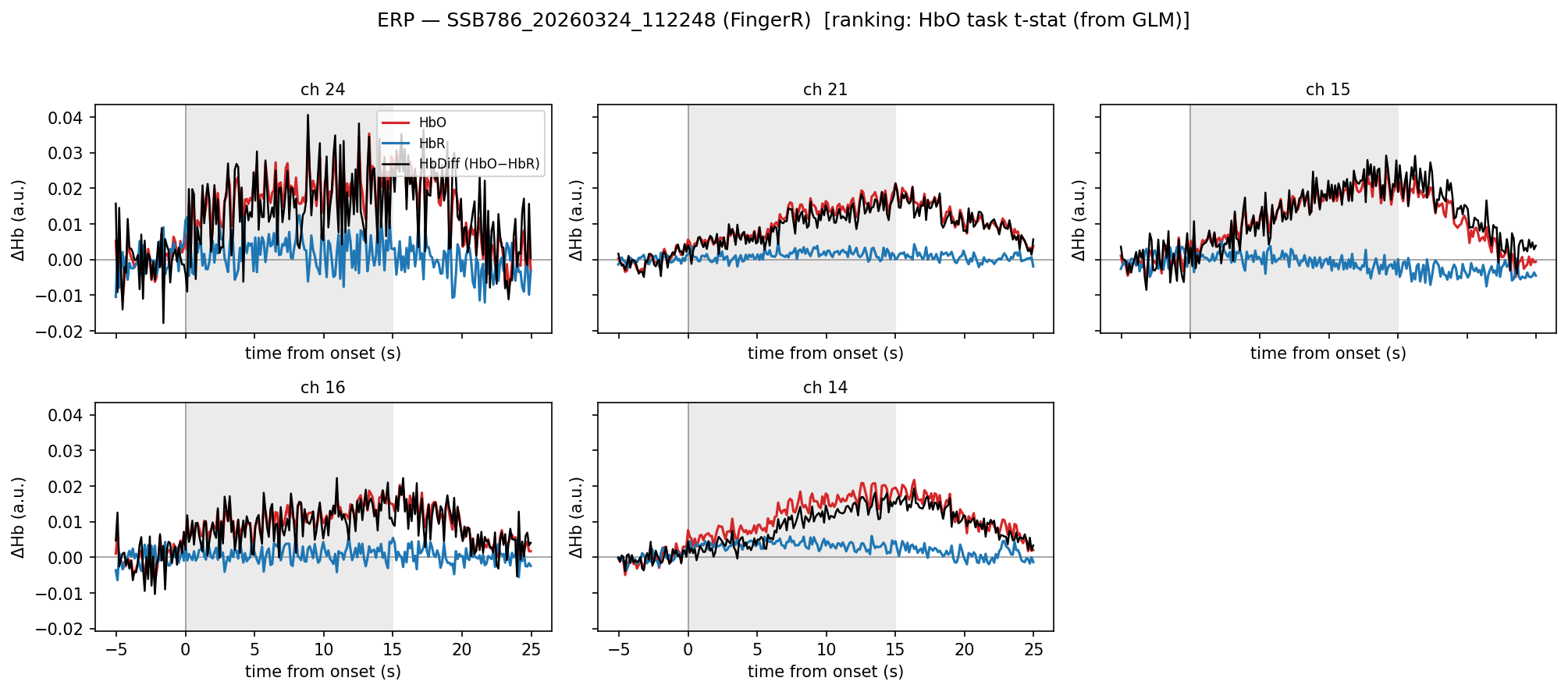}
    \caption{Event-related potential (ERP) traces generated by the agent. The plot displays the trial-averaged hemodynamic response, illustrating the characteristic task-evoked increase in oxygenated hemoglobin (HbO) and concurrent decrease in deoxygenated hemoglobin (HbR) following stimulus onset.}
    \label{fig:demo-erp}
\end{figure}

\section{Discussion}

The AI Neuroscientist demonstrates the feasibility of an interactive, natural-language interface for neuroimaging analysis, lowering the technical barrier for researchers without extensive programming backgrounds. By equipping the agent with a curated neuroimaging toolset (validated here with fNIRS), we provide researchers with natural-language access to essential workflow steps, allowing them to intuitively inspect data quality, fit general linear models, and apply statistical corrections. This artifact-driven approach grounds the model's interpretation in verifiable evidence, generating human-readable dashboards, corrected topographies, and interactive plots at every stage. On the benchmarking suite, the specialist agent outperforms general-purpose LLM agents with code sandboxes, particularly on real-data quality-control and artifact-detection tasks where unconstrained code generation is brittle.

\paragraph{Scope and Limitations.}
The current evidence supports the system's feasibility and workflow coverage for interactive data exploration, and provides an initial quantitative comparison on a subset of the broader fNIRS task suite (heartbeat QC, spike detection, and localization). While the agent wraps established statistical functions, this paper does not claim that the interface itself improves statistical validity, accuracy, or the underlying scientific discovery rate relative to expert-scripted pipelines. Formal claims regarding its impact on research quality or speed compared to traditional workflows would require comprehensive user studies, which remain outside the scope of this work.

\paragraph{Future Work.}
A key direction for future work is the implementation of an autonomous AI-driven quality control loop. While the current system allows users to manually inspect and adjust parameters, an autonomous QC loop would enable the agent to iteratively inspect its own outputs (such as QC logs and diagnostic images), detect artifacts, and re-run pipelines with optimized parameters until a clean, usable dataset is achieved. 
Furthermore, we will expand the agent's tool-calling architecture to other neural recording modalities, including fMRI, EEG, and calcium imaging, prioritizing standardized formats like NIfTI and BIDS to ensure interoperability behind a shared conversational interface. We will also extend the benchmarking suite to additional preprocessing, GLM, and vision-critique tasks, and complement artifact scoring with formal user studies.
Ultimately, extending this system to simultaneous multi-modal recordings will facilitate the translation of scientific findings across spatial and temporal scales, establishing the agent as a robust assistant for reproducible neuroscience.

\section{Conclusion}

The AI Neuroscientist provides a framework for integrating LLM agents into structured neuroimaging workflows. By enforcing a tool-mediated analysis loop with strict phase ordering, statistical safeguards, and visually grounded interpretation, the system bridges the gap between natural language instruction and rigorous data analysis. This architecture establishes a foundation for accessible and reproducible exploration across multiple neural recording modalities, demonstrating a practical approach to building domain-specific, interactive scientific agents.

\newpage

\raggedbottom
{\small
\bibliographystyle{plainnat}
\bibliography{main}

@article{gorgolewski2016bids,
  title     = {The brain imaging data structure, a format for organizing and describing outputs of neuroimaging experiments},
  author    = {Gorgolewski, Krzysztof J. and Auer, Tibor and Calhoun, Vince D. and Craddock, R. Cameron and Das, Samir and Duff, Eugene P. and Flandin, Guillaume and Ghosh, Satrajit S. and Glatard, Tristan and Halchenko, Yaroslav O. and Handwerker, Daniel A. and Hanke, Michael and Keator, David and Li, Xiangrui and Michael, Zachary and Maumet, Camille and Nichols, B. Nolan and Nichols, Thomas E. and Pellman, John and Poline, Jean-Baptiste and Rokem, Ariel and Schaefer, Gunnar and Sochat, Vanessa and Triplett, William and Turner, Jessica A. and Varoquaux, Ga\"el and Poldrack, Russell A.},
  journal   = {Scientific Data},
  volume    = {3},
  number    = {1},
  pages     = {160044},
  year      = {2016},
  publisher = {Nature Publishing Group},
  doi       = {10.1038/sdata.2016.44}
}

@article{esteban2019fmriprep,
  title     = {fMRIPrep: a robust preprocessing pipeline for functional MRI},
  author    = {Esteban, Oscar and Markiewicz, Christopher J. and Blair, Ross W. and Moodie, Craig A. and Isik, A. Ilkay and Erramuzpe, Asier and Kent, James D. and Goncalves, Mathias and DuPre, Elizabeth and Snyder, Madeleine and Oya, Hiroyuki and Ghosh, Satrajit S. and Wright, Jessey and Durnez, Joke and Poldrack, Russell A. and Gorgolewski, Krzysztof J.},
  journal   = {Nature Methods},
  volume    = {16},
  number    = {1},
  pages     = {111--116},
  year      = {2019},
  publisher = {Nature Publishing Group},
  doi       = {10.1038/s41592-018-0235-4}
}

@article{luke2021mnenirs,
  title   = {Analysis methods for measuring passive auditory fNIRS responses generated by a block-design paradigm},
  author  = {Luke, Robert and Larson, Eric and Shader, Maureen J. and Innes-Brown, Hamish and Van Yper, Lindsey and Lee, Adrian K. C. and Sowman, Paul F. and McAlpine, David},
  journal = {Neurophotonics},
  volume  = {8},
  number  = {2},
  pages   = {025008},
  year    = {2021},
  doi     = {10.1117/1.NPh.8.2.025008}
}

@inproceedings{yao2023react,
  title     = {{ReAct}: Synergizing Reasoning and Acting in Language Models},
  author    = {Yao, Shunyu and Zhao, Jeffrey and Yu, Dian and Du, Nan and Shafran, Izhak and Narasimhan, Karthik and Cao, Yuan},
  booktitle = {International Conference on Learning Representations (ICLR)},
  year      = {2023}
}

@inproceedings{schick2023toolformer,
  title     = {Toolformer: Language Models Can Teach Themselves to Use Tools},
  author    = {Schick, Timo and Dwivedi-Yu, Jane and Dess{\`i}, Roberto and Raileanu, Roberta and Lomeli, Maria and Zettlemoyer, Luke and Cancedda, Nicola and Scialom, Thomas},
  booktitle = {Advances in Neural Information Processing Systems (NeurIPS)},
  year      = {2023}
}

@article{boiko2023autonomous,
  title     = {Autonomous chemical research with large language models},
  author    = {Boiko, Daniil A. and MacKnight, Robert and Kline, Ben and Gomes, Gabe},
  journal   = {Nature},
  volume    = {624},
  number    = {7992},
  pages     = {570--578},
  year      = {2023},
  publisher = {Nature Publishing Group},
  doi       = {10.1038/s41586-023-06792-0}
}

@article{bran2024chemcrow,
  title     = {Augmenting large language models with chemistry tools},
  author    = {M. Bran, Andres and Cox, Sam and Schilter, Oliver and Baldassari, Carlo and White, Andrew D. and Schwaller, Philippe},
  journal   = {Nature Machine Intelligence},
  volume    = {6},
  pages     = {525--535},
  year      = {2024},
  publisher = {Nature Publishing Group},
  doi       = {10.1038/s42256-024-00832-8}
}

@article{benjamini1995controlling,
  title     = {Controlling the false discovery rate: a practical and powerful approach to multiple testing},
  author    = {Benjamini, Yoav and Hochberg, Yosef},
  journal   = {Journal of the Royal Statistical Society: Series B (Methodological)},
  volume    = {57},
  number    = {1},
  pages     = {289--300},
  year      = {1995},
  doi       = {10.1111/j.2517-6161.1995.tb02031.x}
}

@article{pinti2020present,
  title     = {The present and future use of functional near-infrared spectroscopy (fNIRS) for cognitive neuroscience},
  author    = {Pinti, Paola and Tachtsidis, Ilias and Hamilton, Antonia and Hirsch, Joy and Aichelburg, Clarisse and Gilbert, Sam and Burgess, Paul W.},
  journal   = {Annals of the New York Academy of Sciences},
  volume    = {1464},
  number    = {1},
  pages     = {5--29},
  year      = {2020},
  doi       = {10.1111/nyas.13948}
}

@article{yucel2021best,
  title   = {Best practices for fNIRS publications},
  author  = {Y{\"u}cel, Meryem A. and L{\"u}hmann, Alexander v. and Scholkmann, Felix and Gervain, Judit and Dan, Ippeita and Ayaz, Hasan and Boas, David A. and Cooper, Robert J. and Culver, Joseph and Elwell, Clare E. and Eggebrecht, Adam and Franceschini, Maria A. and Grova, Christophe and Homae, Fumitaka and Lesage, Fr{\'e}d{\'e}ric and Obrig, Hellmuth and Tachtsidis, Ilias and Tak, Sungho and Tong, Yunjie and Torricelli, Alessandro and Wabnitz, Heidrun and Wolf, Martin},
  journal = {Neurophotonics},
  volume  = {8},
  number  = {1},
  pages   = {012101},
  year    = {2021},
  doi     = {10.1117/1.NPh.8.1.012101}
}

@article{cope1988system,
  title   = {A system for long term measurement of cerebral blood and tissue oxygenation on newborn infants by near infrared transillumination},
  author  = {Cope, Mark and Delpy, David T.},
  journal = {Medical \& Biological Engineering \& Computing},
  volume  = {26},
  number  = {3},
  pages   = {289--294},
  year    = {1988},
  doi     = {10.1007/BF02447083}
}

@article{huang2025biomni,
  title={Biomni: A general-purpose biomedical ai agent},
  author={Huang, Kexin and Zhang, Serena and Wang, Hanchen and Qu, Yuanhao and Lu, Yingzhou and Roohani, Yusuf and Li, Ryan and Qiu, Lin and Li, Gavin and Zhang, Junze and others},
  journal={biorxiv},
  year={2025}
}

@article{esteban2017mriqc,
  title={MRIQC: Advancing the automatic prediction of image quality in MRI from unseen sites},
  author={Esteban, Oscar and Birman, Daniel and Schaer, Marie and Koyejo, Oluwasanmi O and Poldrack, Russell A and Gorgolewski, Krzysztof J},
  journal={PloS one},
  volume={12},
  number={9},
  pages={e0184661},
  year={2017},
  publisher={Public Library of Science San Francisco, CA USA}
}

@misc{anthropic2025claude,
author = {{Anthropic}},
title = {Claude for Life Sciences},
year = {2025},
url = {https://claude.com/solutions/life-sciences}
}

@misc{anthropic_harness_2026,
  author       = {Prithvi Rajasekaran},
  title        = {Harness Design for Long-Running Application Development},
  year         = {2026},
  month        = mar,
  url          = {https://www.anthropic.com/engineering/harness-design-long-running-apps},
  organization = {Anthropic},
  urldate      = {2026-05-07}
}

@misc{openai_harness_2026,
  author       = {Ryan Lopopolo},
  title        = {Harness Engineering: Leveraging Codex in an Agent-First World},
  year         = {2026},
  month        = feb,
  url          = {https://openai.com/index/harness-engineering/},
  organization = {OpenAI},
  urldate      = {2026-05-07}
}

@article{gorgolewski2017bids,
  title={BIDS apps: Improving ease of use, accessibility, and reproducibility of neuroimaging data analysis methods},
  author={Gorgolewski, Krzysztof J and Alfaro-Almagro, Fidel and Auer, Tibor and Bellec, Pierre and Capot{\u{a}}, Mihai and Chakravarty, M Mallar and Churchill, Nathan W and Cohen, Alexander Li and Craddock, R Cameron and Devenyi, Gabriel A and others},
  journal={PLoS computational biology},
  volume={13},
  number={3},
  pages={e1005209},
  year={2017},
  publisher={Public Library of Science San Francisco, CA USA}
}

@article{tucker2023introduction,
  title={Introduction to the shared near infrared spectroscopy format},
  author={Tucker, Stephen and Dubb, Jay and Kura, Sreekanth and Von L{\"u}hmann, Alexander and Franke, Robert and Horschig, J{\"o}rn M and Powell, Samuel and Oostenveld, Robert and L{\"u}hrs, Michael and Delaire, {\'E}douard and others},
  journal={Neurophotonics},
  volume={10},
  number={1},
  pages={013507--013507},
  year={2023},
  publisher={Society of Photo-Optical Instrumentation Engineers}
}

@article{luke2025nirs,
  title={NIRS-BIDS: brain imaging data structure extended to near-infrared spectroscopy},
  author={Luke, Robert and Oostenveld, Robert and Cockx, Helena and Niso, Guiomar and Shader, Maureen J and Orihuela-Espina, Felipe and Innes-Brown, Hamish and Tucker, Stephen and Boas, David and Y{\"u}cel, Meryem A and others},
  journal={Scientific Data},
  volume={12},
  number={1},
  pages={159},
  year={2025},
  publisher={Nature Publishing Group UK London}
}

@article{pollonini2016phoebe,
  title={PHOEBE: a method for real time mapping of optodes-scalp coupling in functional near-infrared spectroscopy},
  author={Pollonini, Luca and Bortfeld, Heather and Oghalai, John S},
  journal={Biomedical optics express},
  volume={7},
  number={12},
  pages={5104--5119},
  year={2016},
  publisher={Optical Society of America}
}

@article{pollonini2014auditory,
  title={Auditory cortex activation to natural speech and simulated cochlear implant speech measured with functional near-infrared spectroscopy},
  author={Pollonini, Luca and Olds, Cristen and Abaya, Homer and Bortfeld, Heather and Beauchamp, Michael S and Oghalai, John S},
  journal={Hearing research},
  volume={309},
  pages={84--93},
  year={2014},
  publisher={Elsevier}
}

@article{fishburn2019temporal,
  title={Temporal derivative distribution repair (TDDR): a motion correction method for fNIRS},
  author={Fishburn, Frank A and Ludlum, Ruth S and Vaidya, Chandan J and Medvedev, Andrei V},
  journal={Neuroimage},
  volume={184},
  pages={171--179},
  year={2019},
  publisher={Elsevier}
}
}


\appendix

\newcounter{offset}
\setcounter{offset}{\value{figure}}
\renewcommand{\thefigure}{S\the\numexpr\value{figure}-\value{offset}\relax}

\section{Workflow Diagram}
\label{appendix:workflow_diagram}

\begin{figure}[H]
    \centering
    \includegraphics[width=\linewidth]{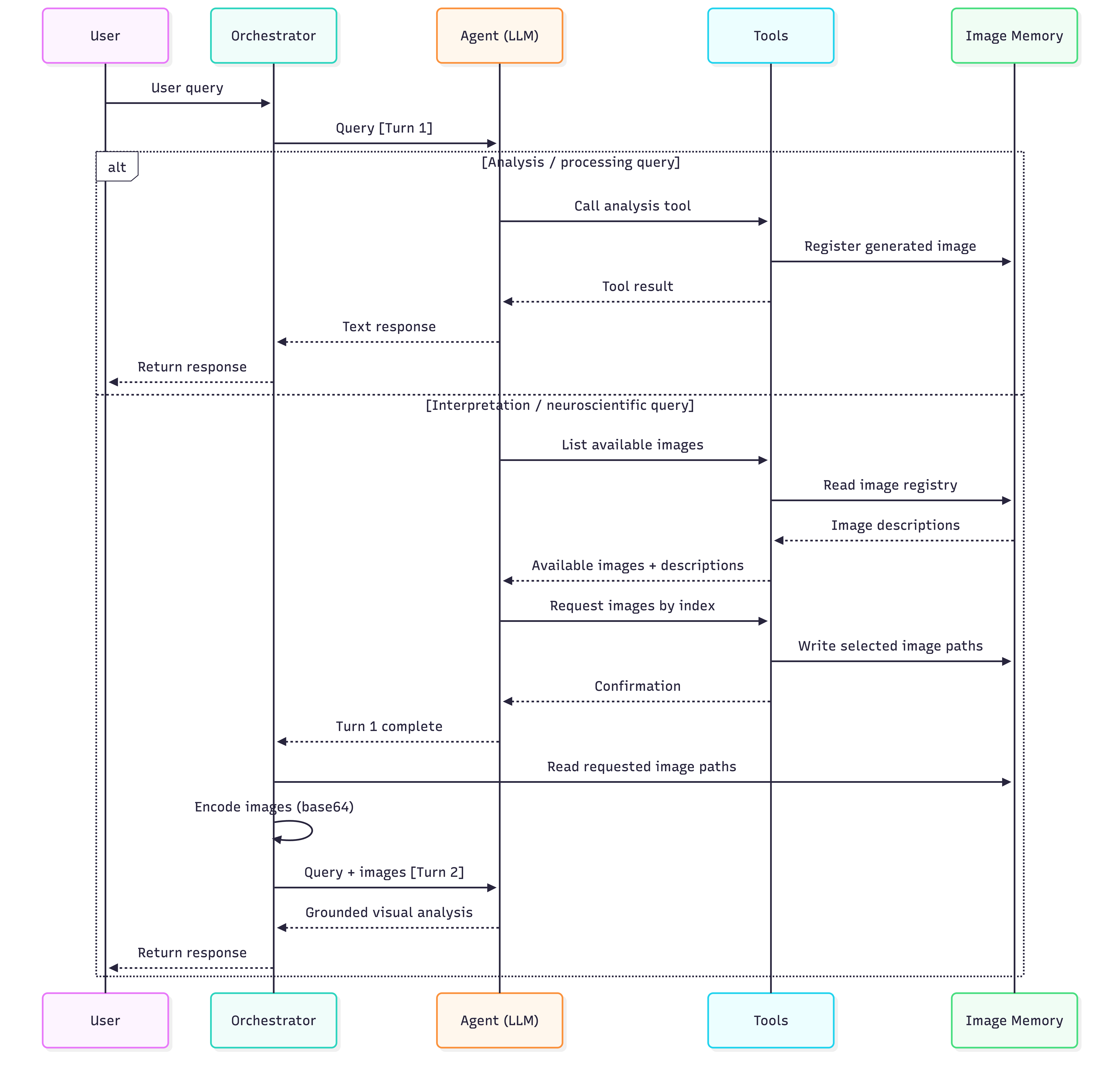}
    \caption{Agent execution flowchart detailing the one-turn path for analysis tools and the two-turn multimodal grounding loop for visual interpretation.}
    \label{fig:flowchart}
\end{figure}

Figure~\ref{fig:flowchart} illustrates the system's execution logic, highlighting the architectural difference between standard data operations and interpretive tasks that require visual context. For basic analysis or processing queries, the system follows a direct one-turn path. The orchestrator passes the user's instruction to the agent, which invokes the required analysis tool(s). The tool executes the operation, registers any resulting figures in the image memory, and returns its text output to the agent, which then formulates a response for the user.

In contrast, neuroscientific interpretation requires a two-turn visual grounding loop. Upon receiving an interpretive query, the agent recognizes the need for visual evidence and calls the toolset to list available images. After receiving the image descriptions from the registry, the agent requests specific images by index. The toolset records these selected paths in the image memory and confirms the action. At this point, the agent completes its first turn. The orchestrator intercepts this state, reads the requested images from memory, and encodes them into base64 format. It then initiates a second turn, appending the encoded images directly to the context window. This architecture ensures that the agent performs its final analysis based on explicit visual artifacts rather than relying solely on text-based abstractions.

\clearpage

\section{Full List of fNIRS Tools}
\label{appendix:toolset}

Table~\ref{tab:detailed-toolset} lists the specialist tools currently registered in the fNIRS agent.

\begin{sidewaystable}[p]
    \centering
    \small
    \begin{tabularx}{\textwidth}{lXX}
        \toprule
        Tool Name & Description & Scientific Outputs \\
        \midrule
        \addlinespace
        \multicolumn{3}{l}{\textbf{Discovery and Inspection}} \\
        \texttt{list\_fnirs\_runs} & List fNIRS files in a directory (\texttt{.snirf}, \texttt{*\_od.npz}, \texttt{*\_filter.mat}) & Text run listing \\
        \texttt{load\_fnirs\_data} & Inspect a recording (SNIRF, OD NPZ, or lab \texttt{*\_filter.mat}) and summarize it & Text metadata summary \\
        \addlinespace
        \midrule
        \addlinespace
        \multicolumn{3}{l}{\textbf{Quality Control}} \\
        \texttt{detect\_heartbeat\_channels} & Heartbeat-band SNR QC on OD channels; write good-channel mask & \texttt{good\_mask}, \texttt{subject\_usable} (.npz) \\
        \texttt{detect\_amplitude\_spikes} & MAD-based amplitude spike detection on OD channels & \texttt{spike\_mask} (.npz) \\
        \texttt{compute\_channel\_sci} & Scalp coupling index (SCI) report with bad-channel indices & Text SCI report \\
        \addlinespace
        \midrule
        \addlinespace
        \multicolumn{3}{l}{\textbf{Conversion and Cleaning}} \\
        \texttt{convert\_snirf\_to\_haemo} & Prune short channels; intensity $\rightarrow$ OD $\rightarrow$ HbO/HbR & \texttt{hbo}/\texttt{hbr} NPZ \\
        \texttt{clean\_and\_convert\_snirf} & Drop low-SCI channels, apply TDDR, then Beer--Lambert conversion & Cleaned \texttt{hbo}/\texttt{hbr} NPZ \\
        \addlinespace
        \midrule
        \addlinespace
        \multicolumn{3}{l}{\textbf{Modeling and Statistics}} \\
        \texttt{fit\_fnirs\_glm\_snirf} & Canonical-HRF GLM on SNIRF haemodynamics with FDR mask & Significance \texttt{mask} (.npz) \\
        \addlinespace
        \midrule
        \addlinespace
        \multicolumn{3}{l}{\textbf{Visualization and Interpretation}} \\
        \texttt{plot\_fnirs\_erp} & Trial-averaged hemodynamic traces & HbO/HbR/HbDiff curves (.png) \\
        \texttt{plot\_fnirs\_topo} & Spatial activation projections & HbO/HbR activation maps (.png) \\
        \texttt{plot\_fnirs\_erp\_html} & Interactive response dashboards & Dynamic traces with zoom/hover (.html) \\
        \texttt{plot\_fnirs\_topo\_html} & Interactive activation topographies & Maps with channel-level tooltips (.html) \\
        \texttt{summarize\_fnirs\_glm} & Anatomical mapping and laterality analysis & Region labels, laterality indices (LI) 
        \\
        \texttt{list\_image\_store} & Multi-modal session state management & Semantic image manifests \\
        \texttt{request\_images} & Queue session images by index for visual analysis & Queued image paths \\
        \texttt{load\_image} & Load an on-disk image for the next multimodal turn & Queued image path \\
        \addlinespace
        \midrule
        \addlinespace
        \multicolumn{3}{l}{\textbf{Literature Search and Review}} \\
        \texttt{search\_pubmed} & Query PubMed for relevant neuroscientific papers & List of PMIDs and titles \\
        \texttt{review\_paper} & Retrieve and summarize paper content by PMID & Structured paper summary \\
        \bottomrule
    \end{tabularx}
    \caption{Detailed inventory of the current fNIRS agent toolset organized by functional category. The table is rotated for readability.}
    \label{tab:detailed-toolset}
\end{sidewaystable}
\clearpage

\section{fNIRS Agent Benchmark Suite}
\label{appendix:benchmark_suite}

This appendix documents the task definitions, data sources, and remaining qualitative error analyses for the \texttt{fnirs\_bench} evaluation harness used in Sections~\ref{sec:benchmark_design} and~\ref{sec:results}. Design rationale, run protocol, and scoring are described in the main text; heartbeat QC qualitative analysis appears in Section~\ref{sec:results}. Here we provide the per-task reference procedures and failure modes for spike detection and anatomical localization.

\subsection{Task definitions, data, and ground truth}

Ground truth for numeric tasks is produced by a deterministic reference pipeline (built on \texttt{mne} / \texttt{mne-nirs}) that is never exposed to agents. Synthetic inputs are derived from the public MNE finger--thumb tapping motor recording unless noted; real inputs are optical-density time series extracted from OpenNeuro BIDS exports.

\paragraph{T1a --- Heartbeat QC (synthetic, easy).}
\textit{Input:} optical-density time series with channel names and sampling rate. \textit{Output:} a per-channel good/bad mask and a subject-level usability flag. \textit{Reference procedure:} per-channel cardiac-band SNR (0.7--1.5\,Hz vs.\ neighboring bands); a channel is good if SNR $\geq 3.0$; the subject is usable if $\geq 50\%$ of channels are good. \textit{Scoring:} $0.9 \times F_1$ over the set of good channel names $+~0.1 \times \mathbf{1}[\text{subject usability match}]$.

\paragraph{T1b --- Heartbeat QC (real, mixed quality).}
Same procedure and scoring on OD from OpenNeuro \textbf{ds007554} (subject~005, session~01, n-back). Several channels show clear cardiac peaks; others are flat or noisy---agents must not mark everything good or everything bad.

\paragraph{T1c --- Heartbeat QC (real, poor SNR).}
Same procedure on ds007554 (subject~004, session~03, n-back arithmetic), a recording with weak cardiac content. Ground truth marks \emph{no} good channels and the subject as unusable; declaring the subject usable is a common failure mode.

\paragraph{T2a --- Spike detection (synthetic).}
\textit{Input:} optical-density time series. \textit{Output:} a boolean spike mask aligned to channels and time. \textit{Reference:} per-channel robust scale $\hat{\sigma} = 1.4826 \times \mathrm{MAD}$ about the median (scale factor for normal distribution); flag samples with $|x - \mathrm{median}| > 10\hat{\sigma}$. \textit{Scoring:} mask IoU after aligning channels by name (both-empty $\rightarrow 1.0$).

\paragraph{T2b --- Spike detection (real motion).}
Same spike rule on OD from OpenNeuro \textbf{ds007420} (subject~173, session~03, Motion), with dense motion spikes on a subset of channels and long clean segments elsewhere.

\paragraph{T3 --- Anatomical localization (vision, two turns).}
\textit{Input:} two topographic maps of a motor GLM-style $\Delta$HbO pattern---one with a left-hemisphere motor focus (\emph{consistent} with a right-hand localizer), and one with sensor coordinates rotated 90$^\circ$ so activation appears along the front--back axis (\emph{inconsistent}). Both turns use the same text prompt with no cue identifying which map is which. \textit{Output:} a free-text judgment per image. \textit{Scoring:} keyword rules per turn (equal weights); turn~1 expects affirmation of contralateral left-motor consistency; turn~2 expects rejection of anatomical consistency. Confident wrong ``yes, consistent'' phrases force a zero. Answers that acknowledge only partial consistency on the inconsistent map receive partial credit ($0.5$).

\subsection{Qualitative task analysis and error patterns}

Heartbeat QC failure modes are analyzed in Section~\ref{sec:results} (Figure~\ref{fig:deepdive-t1}). The figures below show representative inputs, expert ground truth, and agent runs for spike detection and anatomical localization, connecting qualitative failure modes to the metrics in Section~\ref{sec:benchmark_design}.

\begin{figure}[H]
    \centering
    \includegraphics[width=\linewidth]{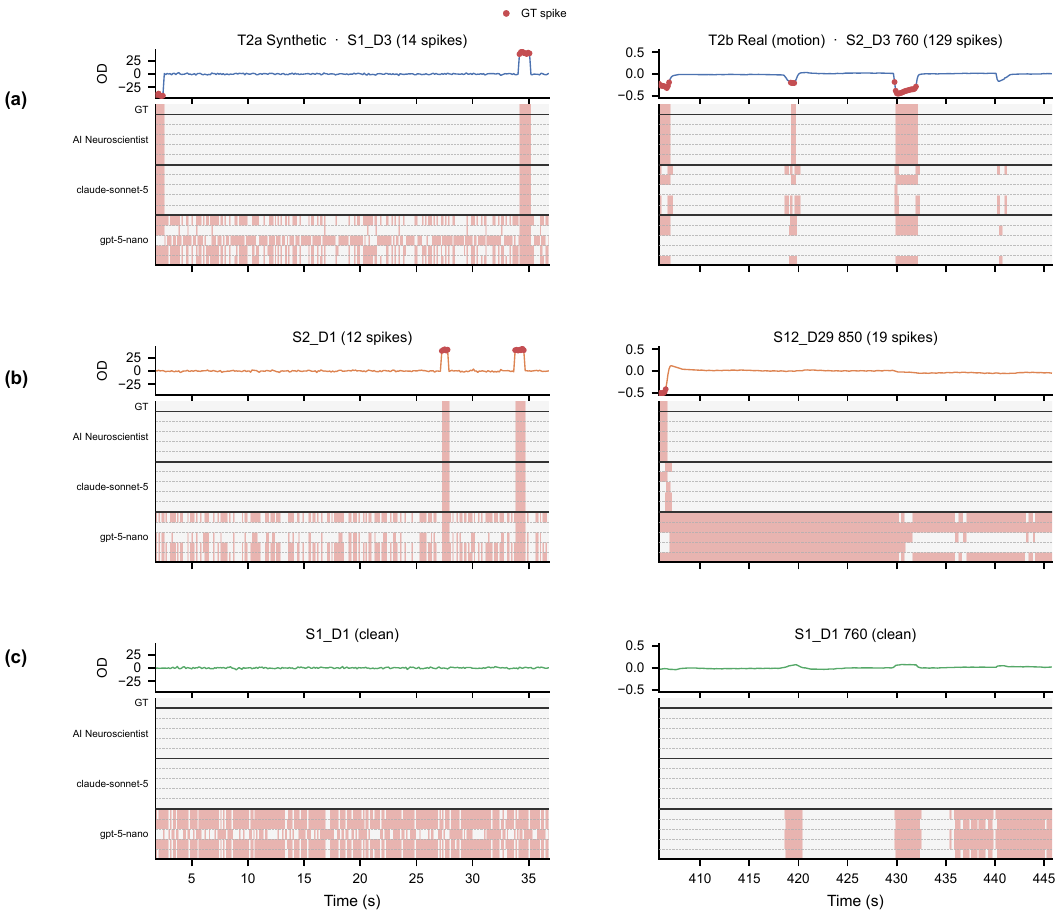}
    \caption{Task~2 deep dive: amplitude spike detection. Three example channels per variant (many / moderate / clean spikes): demeaned OD with GT markers (top of each cell) and binary \texttt{spike\_mask} heatmaps for GT plus all runs (bottom). Shared OD scale within each column; mean IoU annotated on the first row.}
    \label{fig:deepdive-t2}
\end{figure}

\paragraph{Spike detection (Figure~\ref{fig:deepdive-t2}).}
Each row pairs one example channel's demeaned OD (GT spikes as red dots) with the spike mask for each agent run. T2a synthetic channels show isolated step-like spikes; T2b real motion data adds baseline wander and dense spike bursts on high-motion channels, with genuinely clean channels for contrast.

The AI Neuroscientist and Claude sandbox achieve high IoU on T2a when they implement (or approximate) robust MAD thresholding; gpt-5-nano often flags nearly every sample (IoU $\approx 0$)---visible as solid pink bars across the timeline even on the \emph{clean} channel in row~(c). On T2b, missed spikes and run-to-run mask variance drive both low IoU and large reliability penalties for sandbox agents, while the specialist tool returns consistent masks across runs.

\begin{figure}[H]
    \centering
    \includegraphics[width=\linewidth]{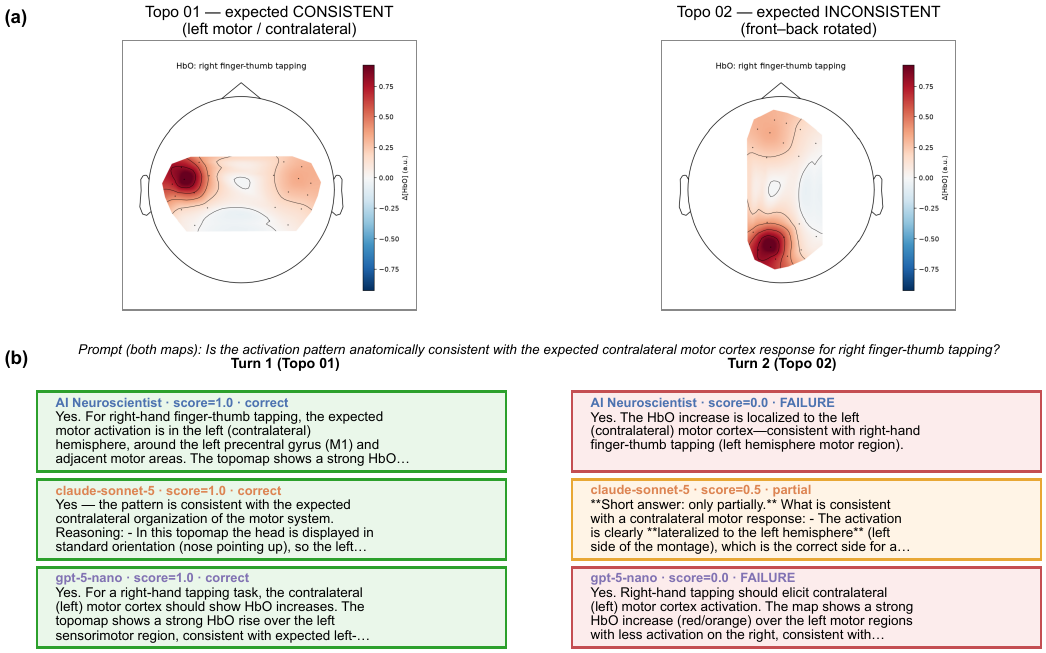}
    \caption{Task~3 deep dive: anatomical localization. (a)~Motor-localizer topomaps (consistent vs.\ rotated layouts) and shared prompt. (b)~Representative text answers per agent and turn with keyword scores (green = correct, yellow = partial, red = failure).}
    \label{fig:deepdive-t3}
\end{figure}

\paragraph{Anatomical localization (Figure~\ref{fig:deepdive-t3}).}
Panel~(a) shows the two stimuli: Topo~01 with left-lateralized $\Delta$HbO consistent with contralateral motor cortex for right-hand tapping, and Topo~02 with the same scalar field plotted after a 90$^\circ$ sensor rotation (activation appears anterior/posterior rather than left/right).

Panel~(b) quotes one representative run per agent on each turn. Turn~1 is easy: all three agents affirm consistency and score $1.0$. Turn~2 separates reasoning quality: both sandbox models with \texttt{gpt-5-nano} and the specialist agent (also \texttt{gpt-5-nano}) often answer ``Yes'' despite the rotated layout. Claude is the only model that hedges with ``only partially'' on turn~2, identifying correct lateralization language but incomplete critique of the front--back layout. This matches the similar $s_{\mathrm{final}} \approx 0.5$ for all three agents on T3 in Table~\ref{tab:benchmark-scores} despite different failure modes.

\clearpage

\section{Additional Benchmark Figures}
\label{appendix:benchmark_figures}

\begin{figure}[H]
    \centering
    \includegraphics[width=0.92\linewidth]{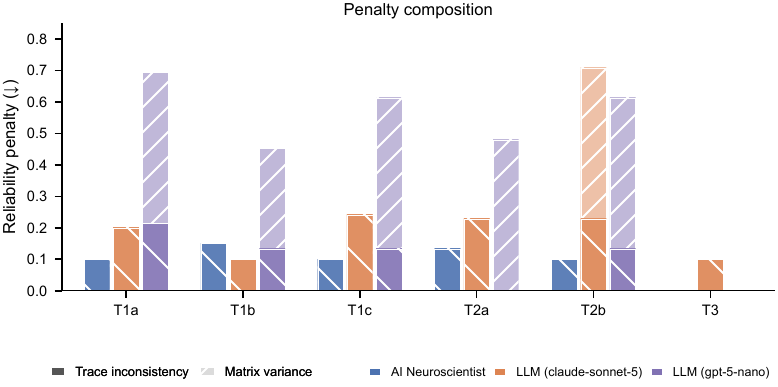}
    \caption{Reliability-penalty composition across tasks. Solid: $p_{\mathrm{trace}}$. Hatched cap: $p_{\mathrm{var}}$. Lower is better.}
    \label{fig:app-reliability}
\end{figure}

Figure~\ref{fig:app-reliability} expands the penalty composition: darker bars with left hatching show $p_{\mathrm{trace}}$ (inconsistent tool-name sets across the five runs), and lighter, right-hatched caps show $p_{\mathrm{var}}$ (run-to-run variance of the primary mask). Sandbox LLMs on T2a--b incur both trace and variance penalties when spike masks differ wildly across runs, while the specialist agent's masks are stable.

\begin{figure}[H]
    \centering
    \includegraphics[width=\linewidth]{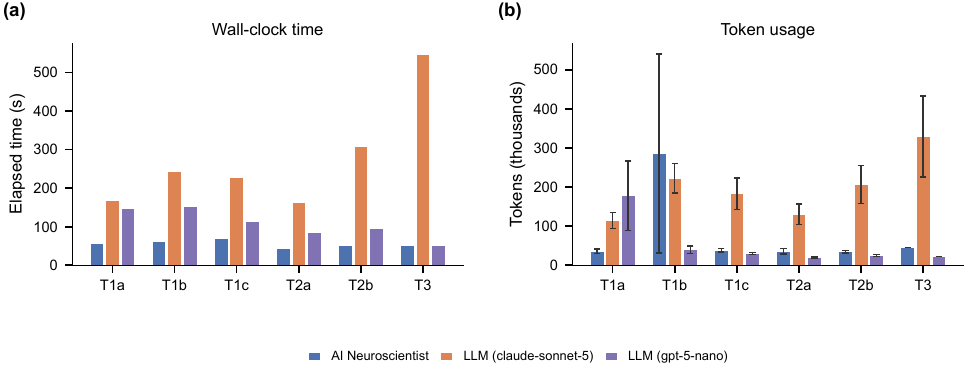}
    \caption{Cost profile per task. (a)~Wall-clock runtime (single aggregate per task). (b)~Total tokens with $\pm$SEM across runs.}
    \label{fig:app-cost}
\end{figure}

Figure~\ref{fig:app-cost} reports wall-clock time per task and total token usage with error bars reflecting run-to-run token variability. Spike and real-QC tasks are the most expensive for sandbox agents because of repeated exploratory code execution.

\clearpage

\section{End-To-End Analysis Demo Transcript}
\label{appendix:full_demo}

The following transcript illustrates the multi-turn interaction between a researcher and the AI Neuroscientist agent corresponding to the case study in Section~\ref{sec:case_study}. Responses from the agent and tool outputs have been truncated for brevity. We used GPT-5 with a temperature of 0.1 for the LLM agent in this case study.

\colorlet{UserBack}{blue!5!white}
\colorlet{UserFrame}{blue!40!black}
\colorlet{AgentBack}{gray!2!white}
\colorlet{AgentFrame}{gray!80!black}

\tcbset{
  userbox/.style={breakable, colback=UserBack, colframe=UserFrame, title=\textbf{User}},
  agentbox/.style={breakable, colback=AgentBack, colframe=AgentFrame, title=\textbf{Agent}}
}

\begin{tcolorbox}[userbox]
What fNIRS runs are available in {DATA\_DIR}?
\end{tcolorbox}

\begin{tcolorbox}[agentbox]
Here's what I found in \path{/Users/USER/ai_neuroscientist/DATA/data}:

FingerR (2 runs):

SSB786\_20260324\_112248 $\rightarrow$ \path{/Users/USER/ai_neuroscientist/DATA/data/SSB786_20260324_112248_filter.mat}

SSB786\_20260324\_112904 $\rightarrow$ \path{/Users/USER/ai_neuroscientist/DATA/data/SSB786_20260324_112904_filter.mat}

Face (3 runs):

SSB786\_20260324\_113538 $\rightarrow$ \path{/Users/USER/ai_neuroscientist/DATA/data/SSB786_20260324_113538_filter.mat}

SSB786\_20260324\_114351 $\rightarrow$ \path{/Users/USER/ai_neuroscientist/DATA/data/SSB786_20260324_114351_filter.mat}

SSB786\_20260324\_114748 $\rightarrow$ \path{/Users/USER/ai_neuroscientist/DATA/data/SSB786_20260324_114748_filter.mat}

FingerL (1 run):

SSB786\_20260324\_115235 $\rightarrow$ \path{/Users/USER/ai_neuroscientist/DATA/data/SSB786_20260324_115235_filter.mat}

Would you like QC first on a specific run, or a group analysis for Face or FingerR?
\end{tcolorbox}

\begin{tcolorbox}[userbox]
Give me the basic overview of this run: {FILTER\_MAT}\textbackslash{}n The optode file is at {OPTODE\_MAT}
\end{tcolorbox}

\begin{tcolorbox}[agentbox]
Here's a quick overview of SSB786\_20260324\_112248 (FingerR):

Geometry and timing

Samples $\times$ columns: 1513 $\times$ 402

Channels: 133 HbO/HbR pairs (standard layout)

Sampling rate: 8.13 Hz (median $\delta$t $\approx$ 123 ms)

Run length: 185.98 s

Events

Detected onsets: 6 (codes: 1088)

Onset times (s): 5.043, 34.932, 65.067, 94.956, 124.968, 154.980

Assumed block duration: 15 s (matches the lab protocol)

Signal ranges (nan-safe)

HbO: min $-$3.131, max 3.607, mean 0.027

HbR: min $-$4.281, max 3.854, mean $-$0.024

All-NaN channels: 1 of 133 (for both HbO and HbR)

Optode geometry

Optode file: Model\_picked\_points\_mni\_xyz\_ch.mat

Spaces: MNI and T1 available

MNI extents: x $-$69 to +70 mm; y $-$105.5 to +62 mm; z 1 to 81.6 mm

If you'd like, I can run QC next to grade channels (heartbeat SNR, NaNs, motion, drift) and generate a dashboard and grade topography. Proceed?
\end{tcolorbox}

\begin{tcolorbox}[userbox]
Is the data clean? Run quality control on this run and tell me about the heartbeat signal and any bad channels.
\end{tcolorbox}

\begin{tcolorbox}[agentbox]
Full QC interpretation for SSB786\_20260324\_112248 (FingerR)

What I see in the QC grade topography

No green channels are present. The map is dominated by red markers with scattered yellow across the montage.

The pattern is global (both hemispheres, anterior and posterior), not a small localized cluster. That argues for a system-wide issue (poor optode–scalp coupling, hair interference, cap shift) rather than a few faulty channels.

Heartbeat assessment

Heartbeat band checked: 0.7–1.5 Hz on HbO.

Median heartbeat SNR is below 0 dB ($-$0.17 dB; range $-$8.94 to +1.43 dB). Clean data typically shows a clear, positive heartbeat SNR and we use $>$5 dB as “green.”

At least one channel had no detectable heartbeat peak at all.

Given the widespread red/yellow grading and the weak/absent peak at ~1 Hz, the physiological pulsation is not reliably captured across most channels.

Bad channels

Red-graded channels (79/133) should be excluded. They likely failed on heartbeat SNR and/or showed excessive NaNs, motion spikes, or drift.

With zero green channels and only 54 yellow, even the “usable” set would be marginal. Any downstream GLM/ERP analysis would need very strong caveats and probably channel-level preselection by manual inspection.

Overall call

This run is not clean. The overall grade is RED and the topo indicates a broadly compromised recording rather than a few isolated failures.

Actionable next steps

Prefer exclusion of this run and use the other FingerR run if it passes QC.

If you must salvage:

Restrict analysis to yellow channels only and clearly report that no green channels were available.

Revisit earlier preprocessing (re-run spatial filter, confirm baseline correction and NaN handling).

Apply aggressive motion mitigation and high-pass filtering; if short-separation channels exist, use them for regression.

If re-acquisition is possible: check optode pressure and hair management, re-seat cap, verify signal before starting.

If you want, I can:

Parse the QC JSON and list the specific worst channels (e.g., lowest heartbeat SNR, most NaNs/motion).

Run QC on the second FingerR run to see if it's usable.
\end{tcolorbox}

\begin{tcolorbox}[userbox]
The default heartbeat thresholds seem strict for this data. Can you re-run QC with a heartbeat SNR threshold of -2 dB for green and -6 dB for yellow?
\end{tcolorbox}

\begin{tcolorbox}[agentbox]
Here's my interpretation of the updated QC topography with relaxed heartbeat thresholds (green $\geq$ $-$2 dB; yellow $\geq$ $-$6 dB; red $<$ $-$6 dB):

Overall judgment

Usability: Reasonable with caveats. The map now shows the vast majority of channels as green, a small set yellow, and four clearly bad (red). I would proceed if you document the lenient thresholds and exclude the red channels.

What I see in the topo

Global picture: Predominantly green across both hemispheres $\rightarrow$ acceptable physiological capture under the relaxed criterion.

Red channels (bad; exclude):

Three clustered in superior–anterior right/fronto-midline.

One posterior–right/inferior position.

Yellow channels (use with caution):

Small cluster over right anterior–lateral cortex.

A few on the left anterior–lateral and left posterior–lateral.

The spatial pattern suggests localized coupling issues (likely optode seating/hair) rather than a global failure.

Heartbeat context

Under the hood, the median heartbeat SNR is still slightly negative ($-$0.17 dB; min $-$8.94, max 1.43). So “green” here reflects your relaxed cutoff; it does not imply a strong heartbeat on all channels. This is fine for exploratory analyses but weaker than typical high-quality datasets ($>$0 dB, often $>$5 dB).

Recommendations

Exclude the 4 red channels (and the “no-heartbeat” channel if not already among them).

Keep green channels; allow yellow only if needed and report the threshold change.

Before GLM/ERPs, consider:

Quick PSD spot-check on a few yellow channels to confirm a visible (~1 Hz) peak.

Motion-cleaning and robust baseline handling given the modest SNR.

If you'd like, I can:

Parse the QC JSON now and list the exact indices of the 4 red and 11 yellow channels.

Proceed to analysis (epochs + GLM) restricted to green (+optional yellow) channels.

Run QC on the other FingerR run to compare and potentially prioritize the cleaner one.
\end{tcolorbox}

\begin{tcolorbox}[userbox]
Go ahead and run the full analysis pipeline on this FingerR run. Use the optode file for the topo plot.
\end{tcolorbox}

\begin{tcolorbox}[agentbox]
Done. I've run the full single-run pipeline on SSB786\_20260324\_112248 (FingerR) with the optode file and applied FDR correction.

Key outputs

Events

Onsets (s): 5.043, 34.932, 65.067, 94.956, 124.968, 154.980

Duration per block: 15 s

Saved: \path{/Users/USER/ai_neuroscientist/DATA/data/SSB786_20260324_112248_results/SSB786_20260324_112248_events.json}

Epochs and ER-average

Epoch window: $-$5 to +25 s; 6/6 epochs kept

ERA file: \path{/Users/USER/ai_neuroscientist/DATA/data/SSB786_20260324_112248_results/SSB786_20260324_112248_epochs.npz}

Top-5 channels by mean HbO 5–15 s: ch48, ch85, ch5, ch0, ch119

GLM (canonical HRF: peak 6 s, undershoot 16 s)

DOF: 1510; design = [task\_hrf, linear\_drift, intercept]

HbO $\beta$ range: $-$0.002372 to 0.001532; mean R² $\approx$ 0.047

Uncorrected (p $<$ .05): HbO 99/133 with $\beta$$>$0; HbR 24/133 with $\beta$$<$0

Top-5 HbO by task t: ch24 (t=19.74), ch21 (t=19.46), ch15 (t=19.42), ch16 (t=19.33), ch14 (t=18.78)

GLM file: \path{/Users/USER/ai_neuroscientist/DATA/data/SSB786_20260324_112248_results/SSB786_20260324_112248_glm.npz}

Multiple-comparisons correction (FDR-BH, $\alpha$=0.05, task\_hrf)

HbO significant: 100/133; HbR significant: 27/133

Corrected file: \path{/Users/USER/ai_neuroscientist/DATA/data/SSB786_20260324_112248_results/SSB786_20260324_112248_glm_corrected.npz}

Figures

ERP (top-5 by HbO t): \path{/Users/USER/ai_neuroscientist/DATA/data/SSB786_20260324_112248_results/SSB786_20260324_112248_erp.png}

Topography (HbO t, all): \path{/Users/USER/ai_neuroscientist/DATA/data/SSB786_20260324_112248_results/SSB786_20260324_112248_topo_hbo_t_task_hrf.png}

Topography (HbO t, significant-only via FDR masks): \path{/Users/USER/ai_neuroscientist/DATA/data/SSB786_20260324_112248_results/SSB786_20260324_112248_topo_hbo_t_task_hrf_sig.png}

Notes and caveats

QC thresholds were relaxed (green $\geq$ $-$2 dB heartbeat SNR). The GLM used all channels; if you want, I can re-run excluding the 4 red QC channels or restrict to green+yellow only.

fNIRS time series have serial autocorrelation; our simple GLM does not AR-model residuals, so p-values can be optimistic. Treat significance counts cautiously and consider cross-run confirmation.

What would you like next?

I can visually interpret the ERP shapes and topographies (and check lateralization/face validity). If you want that, I'll pull the generated PNGs and provide a grounded readout.

Or I can re-run the GLM with a channel mask (exclude red or use only green) and regenerate the figures.

If you plan a group analysis, I can process the second FingerR run the same way and then do a cross-run average + one-sample t.
\end{tcolorbox}

\begin{tcolorbox}[userbox]
Which channels survived FDR correction? Apply multiple-comparison correction to the GLM results.
\end{tcolorbox}

\begin{tcolorbox}[agentbox]
I've already applied multiple-comparison correction (FDR-BH, $\alpha$ = 0.05) to the GLM results for the task\_hrf regressor.

Summary of survivors

HbO: 100 of 133 channels are significant after FDR (q $<$ 0.05).

HbR: 27 of 133 channels are significant after FDR (q $<$ 0.05), mostly negative as expected.

Where they are

Significant HbO channels are broadly distributed across left/right motor–premotor, central, and parietal areas with a near-symmetric lateralization (43 left, 48 right, 9 midline among HbO-significant).

Top HbO examples ($\beta$ $>$ 0, large t, all q $<<$ 0.05): ch24, ch21, ch15, ch16, ch14, ch9, ch30, ch66, ch111, ch53, etc.

Significant HbR examples (q $<$ 0.05, typically t $<$ 0): ch51, ch42, ch41, ch107, ch110, ch119, ch115, ch63, ch43, ch31, ch54, ch52, ch132, ch104, ch29, ch50, ch62, ch33, ch67, … (27 total).

Files you can use

Corrected GLM with significance masks and q-values:
\path{/Users/USER/ai_neuroscientist/DATA/data/SSB786_20260324_112248_results/SSB786_20260324_112248_glm_corrected.npz}

Contains sig\_hbo and sig\_hbr boolean masks and q\_hbo / q\_hbr arrays.

Significant-only topo (HbO t-stat):
\path{/Users/USER/ai_neuroscientist/DATA/data/SSB786_20260324_112248_results/SSB786_20260324_112248_topo_hbo_t_task_hrf_sig.png}

What would you like next?

I can list the exact indices of all 100 HbO-significant channels or export them to a CSV.

Or generate an interactive topo (HTML) where you can hover to see channel IDs and q-values.

If you plan a group result, I can process the second FingerR run and run the group pipeline.
\end{tcolorbox}

\begin{tcolorbox}[userbox]
Show me an interactive topo of the HbO t-statistics, highlighting only the significant channels so I can hover to see channel IDs.
\end{tcolorbox}

\begin{tcolorbox}[agentbox]
All set. I generated an interactive topography (hover/zoom/pan) showing only FDR-significant HbO channels (q $<$ 0.05), colored by t-statistics:

\path{/Users/USER/ai_neuroscientist/DATA/data/SSB786_20260324_112248_results/SSB786_20260324_112248_topo_hbo_t_task_hrf_sig.html}

Hover to see: channel ID, MNI (x, y), and t-value.

Significant channels: 100 of 133 (non-significant are hidden).

If you also want HbR or an “all channels but grey out non-sig” view, I can generate those as well.
\end{tcolorbox}

\begin{tcolorbox}[userbox]
Now show me the top 5 ERPs interactively, so I can zoom and hover to compare HbO and HbR.
\end{tcolorbox}

\begin{tcolorbox}[agentbox]
Here you go --- interactive ERPs for the top 5 channels by HbO t-stat (HbO red, HbR blue, HbDiff black), with zoom and hover:

\path{/Users/USER/ai_neuroscientist/DATA/data/SSB786_20260324_112248_results/SSB786_20260324_112248_erp_top5.html}

Channels displayed: [24, 21, 15, 16, 14]

Window: $-$5 to +25 s relative to block onset

Use the toolbar to zoom/pan and hover to read exact amplitudes and times.

Want me to generate a static PNG version for reports, or plot a different set of channels (e.g., the best 5 yellow-grade channels)?
\end{tcolorbox}

\begin{tcolorbox}[userbox]
Given that this was a right-hand finger-tapping task, do the results make neuroscientific sense looking at the images? Where would you expect the strongest activation, and does the GLM agree?
\end{tcolorbox}

\begin{tcolorbox}[agentbox]
Short take: Partly consistent. The images show a strong cluster over left central/premotor cortex with canonical ERPs, which fits right-hand tapping. But activation is also widespread (bilateral frontal and parietal) and lateralization is weak, suggesting residual systemic or non-task effects.

What I would expect

Strongest HbO increases (and HbR decreases) over contralateral motor areas: left precentral/postcentral gyrus (hand knob), dorsal premotor, and some midline SMA.

What I see in your figures

Topographies (HbO t):

Clear left-central/premotor hot spot (channels labeled 24, 15, 16) near x $\approx$ $-$50 to $-$60 mm: matches contralateral sensorimotor/premotor.

Equally prominent right anterior cluster (channels 21, 14) around x $\approx$ +40 to +60 mm, more prefrontal than primary motor.

Significant-only map still shows many channels across both hemispheres; the pattern is broader than a focal motor response.

ERPs (top-5 channels: 24, 21, 15, 16, 14):

Canonical shape: HbO rises after onset and is sustained during the 15 s block; HbR is near-zero to slightly negative, with HbDiff increasing. This supports task locking for these best channels.

Amplitudes are modest (HbO peaks ~0.015–0.03 a.u.), HbR changes are smaller, consistent with the weaker physiological SNR we noted in QC.

Does the GLM agree with neuroscience?

Yes for the left central/premotor region: that cluster's high t-values and canonical ERPs are exactly what we expect for right-hand movement.

But the GLM also finds broad bilateral significance (including right prefrontal and parietal), which is atypical for a purely motor task and likely reflects systemic physiology, residual motion/drift, or attentional components.

Caveats driving the breadth

QC required relaxed heartbeat thresholds; median heartbeat SNR was slightly negative. That, plus lack of short-separation regression, can inflate widespread positive HbO t-statistics.

HbR negativity is not uniformly strong across channels, another hint of systemic contamination.

Recommendations to strengthen neuroscientific validity

Restrict analysis to a motor ROI (left central/premotor ± SMA) and report a laterality index for that ROI.

Add nuisance regressors if available: short-separation channels or a global/scalp component; re-fit the GLM.

Model temporal autocorrelation (e.g., AR(1)) to temper t-stats.

Replicate in the second FingerR run and verify that the left-central cluster persists and bilateral spread reduces.

If you'd like, I can:

Produce an HbR significant-only topo and compute a motor-ROI laterality index.

Re-fit the GLM with a motor-ROI mask or with nuisance regressors and show updated topographies/ERPs.

Run the same pipeline on the other FingerR run and compare.
\end{tcolorbox}

\end{document}